\documentclass[letterpaper, 10 pt, conference]{ieeeconf}  

\IEEEoverridecommandlockouts                              

\usepackage{amsmath} 
\usepackage{amssymb}  
\usepackage{algorithm}
\usepackage{algpseudocode}

\usepackage{booktabs}   
\usepackage{multirow}   
\usepackage{multicol}   
\usepackage{array}      
\usepackage{caption}    
\usepackage{graphicx}   
\usepackage[table]{xcolor}  
\usepackage{flushend}
\usepackage{url}
\usepackage{hyperref}
\usepackage{balance}

\usepackage{xcolor}
\usepackage{tcolorbox}
\usepackage{pifont} 

\definecolor{colordraft}{HTML}{FFDD99}    
\definecolor{colorprogress}{HTML}{99CCFF} 
\definecolor{colorready}{HTML}{99FF99}    
\definecolor{colornotstarted}{HTML}{DDDDDD} 

\newcommand{\cmark}{\ding{51}}  
\newcommand{\xmark}{\ding{55}}  

\title{\LARGE \bf
MASAR: Motion--Appearance Synergy Refinement
\\for Joint Detection and Trajectory Forecasting
}

\author{
Mohammed Amine Bencheikh Lehocine$^{1}$,
Julian Schmidt$^{1}$,
Frank Moosmann$^{1}$,
Dikshant Gupta$^{1}$,
and Fabian Flohr$^{2}$%
\thanks{This work is a result of the joint research project STADT:up (19A22006O). The project is supported by the German Federal Ministry for Economic Affairs and Energy (BMWE), based on a decision of the German Bundestag. The author is solely responsible for the content of this publication.}%
\thanks{$^{1}$Mercedes-Benz AG, Germany.}%
\thanks{$^{2}$Munich University of Applied Sciences, Germany.}%
}

\begin{document}

\maketitle
\thispagestyle{empty}
\pagestyle{empty}

\begin{abstract}
Classical autonomous driving systems connect perception and prediction modules via hand-crafted bounding-box interfaces, limiting information flow and propagating errors to downstream tasks. Recent research aims to develop end-to-end models that jointly address perception and prediction; however, they often fail to fully exploit the synergy between appearance and motion cues, relying mainly on short-term visual features. We follow the idea of ``\textit{looking backward to look forward}'', and propose MASAR, a novel fully differentiable framework for joint 3D detection and trajectory forecasting compatible with any transformer-based 3D detector. MASAR employs an object-centric spatio-temporal mechanism that jointly encodes appearance and motion features. By predicting past trajectories and refining them using guidance from appearance cues, MASAR captures long-term temporal dependencies that enhance future trajectory forecasting. Experiments conducted on the nuScenes dataset demonstrate MASAR's effectiveness, showing improvements of over 20\% in minADE and minFDE while maintaining robust detection performance. Code and models are available at
\url{https://github.com/aminmed/MASAR}. 

\end{abstract}

\section{INTRODUCTION}

A fundamental requirement for safe and reliable autonomous driving is the ability to both perceive the surrounding scene and anticipate the future behaviors and interactions of nearby agents~\cite{Xu2024towards}. Vision-based systems have gained increasing attention due to their lower deployment cost, scalability, and rich semantic information. However, camera setups remain highly sensitive to localization errors caused by depth ambiguity and occlusions, which degrade detection performance and, consequently, future prediction. 

To mitigate these issues, prior works~\cite{li2022bevformer, liu2023sparsebev, wang2023exploring, lin2022sparse4d, han2024exploring} leverage historical frames to enhance the robustness of learned scene representations. These methods primarily augment appearance features (high level visual features describing how objects look)  but do not model long-term past object dynamics. Bird’s-eye-view-based (BEV-based) approaches, for instance, compensate for ego-motion when aligning and aggregating past frames~\cite{huang2022bevdet4d, li2022bevformer, han2024exploring,li2024bevnext} but ignore individual object motion. Perspective-based (sparse query-based) models, on the other hand, either focus only on modeling short-term inter-frame object motion ~\cite{wang2023exploring, papais2025foresight, doll2024dualad} or rely on constant-velocity assumptions for temporal features sampling~\cite{liu2023sparsebev, lin2022sparse4d} thereby often overlooking long-term motion patterns that can potentially improve both detection and forecasting.

\begin{figure}[t]
    \centering
    \includegraphics[width=\columnwidth]{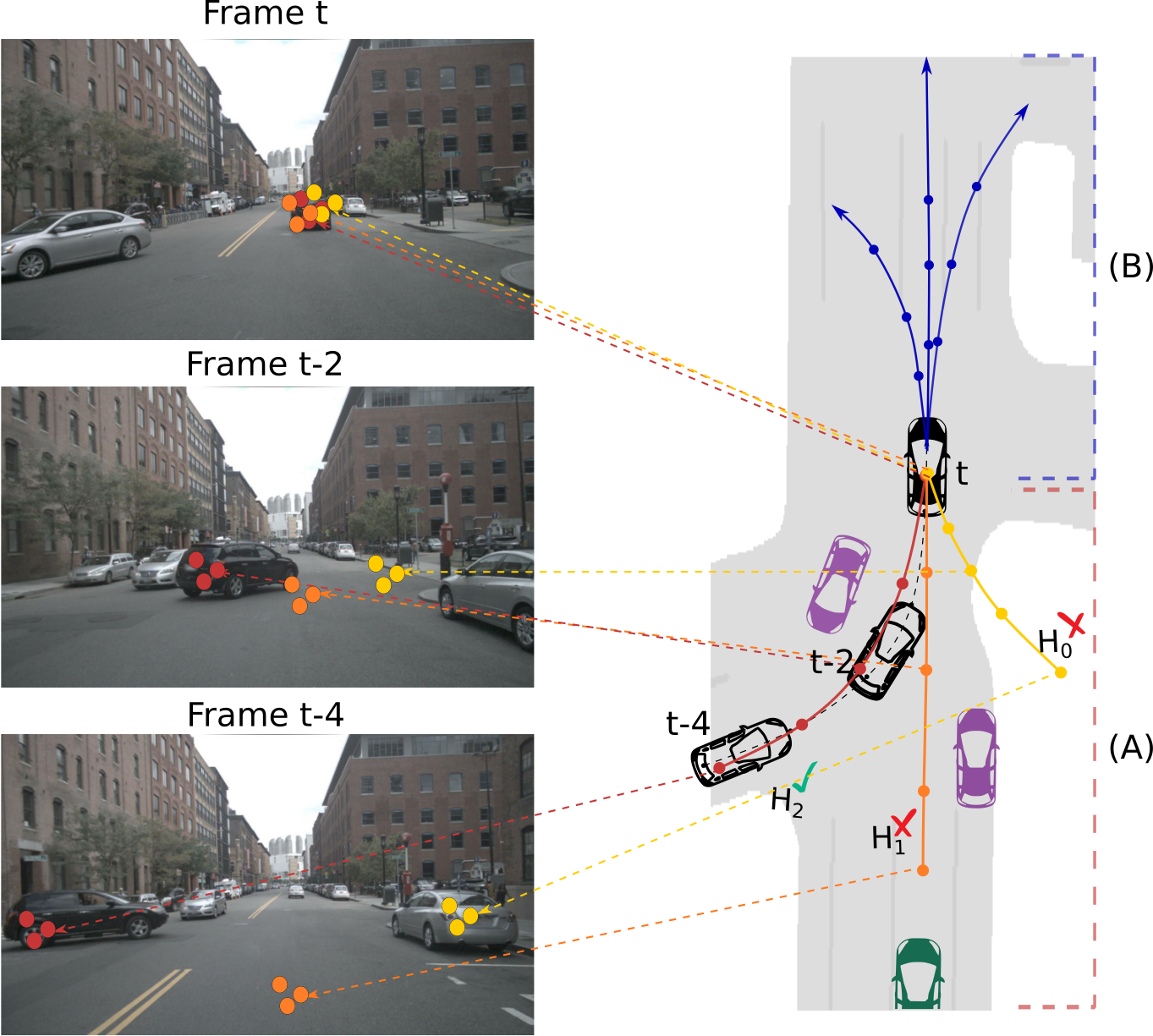} 
    \caption{Core idea of our method. \textit{Left}: multi-frame input images, \textit{Right}: (A) for each object query (i.e. hypothesis), taking the black car as example, we iteratively refine past trajectory hypotheses (yellow, orange, red), aggregate visual features along them, and perform appearance-guided scoring. $\mathbf{H_2}$ wins because it hits more visual features corresponding to the object. (B) Based on the selected past trajectory and aggregated features, multiple modes of future trajectories (blue) are predicted.}
    \label{fig:teaser}
\end{figure}

In modular autonomous driving pipelines, multi-object tracking is used to construct past trajectories for future prediction. Camera-based tracking is inherently noisy~\cite{huang2023delving, Xu2024towards}, prior works~\cite{gu2023vip3d, papais2025foresight} reported that explicitly including noisy tracked trajectories leads to degradation in forecasting performance. Recent end-to-end approaches instead propagate tracking queries as input for forecasting~\cite{gu2023vip3d,hu2023planning,sun2024sparsedrive,papais2025foresight}, but these queries primarily encode re-identification features and do not completely replace the accurate past trajectories.

To overcome these limitations, we bypass noise-prone tracking by directly predicting smooth past trajectories and introduce a new approach that refines past motion while modeling temporal object features. Our framework consists of two key components: the Appearance-guided Past Motion Refinement (APR), which predicts multiple candidate past trajectories and selects the most compatible one using appearance cues, and the Past-conditioned Forecasting Decoder (PFD), which leverages these refined trajectories to improve future prediction (Figure~\ref{fig:teaser}). Unlike prior works~\cite{papais2025foresight,hu2023planning,doll2024dualad} that rely solely on tracking queries to encode object dynamics, PFD explicitly incorporates past trajectories alongside object queries. Notably, MASAR outperforms previous works on end-to-end forecasting without tracking or any map information.

In summary, our contributions are:
\begin{itemize}
    \item We propose MASAR, a tracking-free, map-free framework for joint 3D detection and trajectory forecasting, compatible with any transformer-based detector. To demonstrate its adaptability, we integrate it into two leading architectures: BEVFormer~\cite{li2022bevformer} and SparseBEV~\cite{liu2023sparsebev}\footnote{Despite its name, SparseBEV does not construct a BEV representation; its decoder directly interacts with perspective multi-scale image features.}, achieving consistent improvements.
    \item MASAR sets a new state-of-the-art performance for end-to-end forecasting on the nuScenes dataset~\cite{caesar2020nuscenes}, reducing minADE and minFDE by over 20\% without relying on map information.
    \item Through extensive ablations, we show that conditioning on refined past trajectories provides up to 6\% improvement in minFDE and 7\% reduction in miss rate, highlighting the critical impact of our past-motion modeling approach.
\end{itemize}

\begin{figure*}[t]
    \centering
    \includegraphics[width=1.\textwidth]{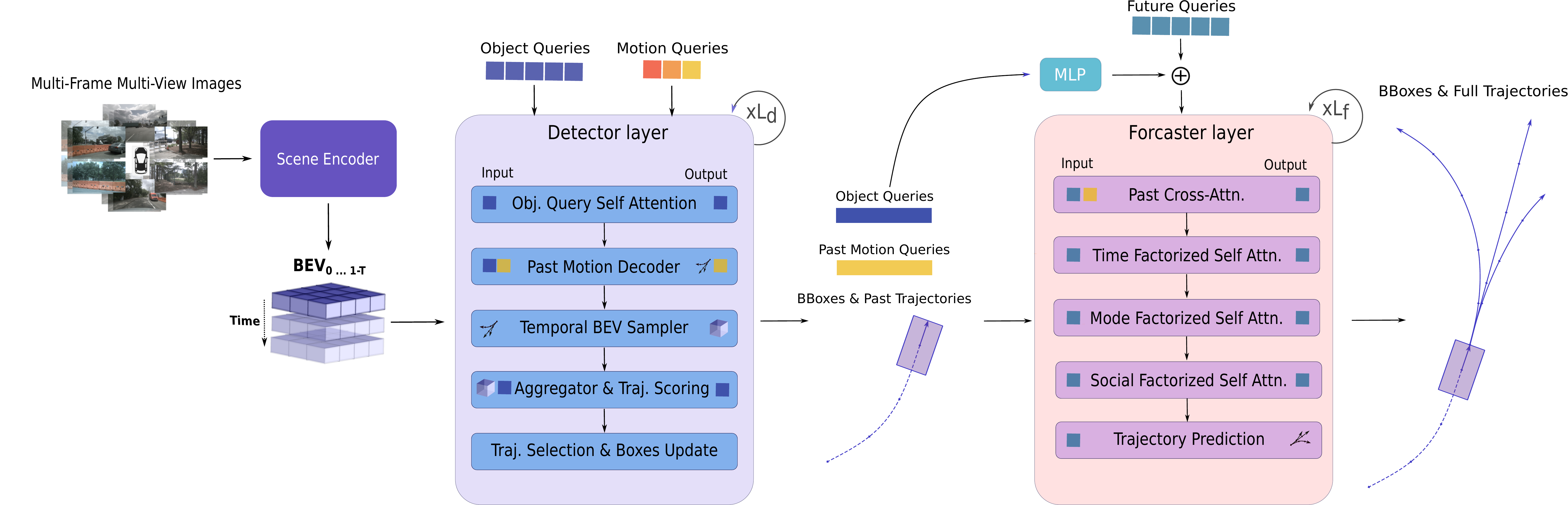} 
    \caption{MASAR architecture: The scene encoder encodes all multi-frame, multi-view images into BEVs. \textit{Detector decoder}: $L_d$ layers iteratively refine object detections together with their past trajectory estimates. \textit{Forecasting decoder}: $L_f$ layers iteratively forecast multiple trajectory modes for each detected object based on its estimated past trajectory and its visual features along that trajectory.}
    \label{fig:model_architecture}
\end{figure*}

\section{RELATED WORKS}

\subsection{Camera-based 3D Object Detection}

Early approaches address multi-view 3D detection by first applying standard 2D methods and then transform the 2D boxes to 3D by regressing additional 3D attributes~\cite{manhardt2019roi}. LSS~\cite{philion2020lss} introduces learning holistic BEV representations via view transformation, either by predicting per-pixel depth distributions to lift 2D features into 3D or BEV space~\cite{reading2021cadnn,huang2022bevdet4d,hu2021fiery}, or by projecting BEV pillars onto the 2D image plane for features' sampling~\cite{li2022bevformer, fang2023tbp}. DETR3D~\cite{wang2022detr3d} leverages sparse object queries with 3D reference points to sample image features. PETR~\cite{liu2022petr} and SpatialDETR~\cite{doll2022spatialdetr} further incorporate 3D geometric priors into image features, while StreamPETR~\cite{wang2023exploring} extends this design to a streaming setting. 

\subsection{Spatio-temporal Modeling} \label{spatio_temporal}
Spatio-temporal modeling is crucial for achieving high performance in camera-based 3D detection. Depending on the intermediate scene representation, two main directions can be distinguished:

\paragraph*{Scene-level Modeling} Often adopted by BEV-based models. Methods such as BEVDet4D~\cite{huang2022bevdet4d}, FIERY~\cite{hu2021fiery}, and BEVerse~\cite{zhang2022beverse} wrap and fuse temporal BEVs using ego-motion compensation and convolutions, while BEVFormer~\cite{li2022bevformer} employs deformable attention to recurrently attend to previous BEVs. VideoBEV~\cite{han2024exploring} uses a hybrid of parallel and recurrent fusion but still relies on wrapping and convolutions. These methods, however, can lose important temporal information due to restricted BEV range and induce feature distortion~\cite{fang2023tbp} because they do not account for individual object motion.

\paragraph*{Object-level Modeling} For perspective-based models, SparseBEV~\cite{liu2023sparsebev} and Sparse4D~\cite{lin2022sparse4d} use a constant velocity model to generate past trajectories for sampling features from previous perspective frames. Inspired by the tracking-by-attention mechanism~\cite{zhang2022mutr3d}, StreamPETR~\cite{wang2023exploring} and Sparse4Dv3~\cite{lin2023sparse4dv3} adopt query propagation for temporal modeling. Additionally, StreamPETR~\cite{wang2023exploring} and~\cite{papais2025foresight, doll2024dualad} incorporate inter-frame ego and object motion using a dedicated motion layer normalization.

A key limitation of prior works is that spatio-temporal modeling focuses on aggregating visual information from past frames without explicitly building up an understanding of objects' past motion. Our APR introduces a novel approach that jointly predicts and refines both long-term past motion and appearance features.

\subsection{Conventional Trajectory Forecasting}
Given HD maps and past agent trajectories, the goal is to predict multiple future trajectories for surrounding agents. Early works rely on rasterized scene representations~\cite{bansal2018chauffeurnet,chai2019multipath}, later~\cite{gao2020vectornet} introduces vectorized representation  to better handle heterogeneous inputs. Recent transformer-based methods~\cite{ngiam2021scene,shi2022motion,zhou2023query} achieve state-of-the-art performance and solve challenging scenarios across multiple datasets. However, the assumption of access to curated past trajectories is often unrealistic in real-world settings~\cite{Xu2024towards}, thereby limiting the applicability of these methods in practical autonomous driving scenarios. 

\subsection{Joint Detection and Forecasting} 

Detection and forecasting directly from sensor inputs has recently gained significant attention. Early LIDAR-based approaches employ convolutional networks to jointly detect, track, and forecast motion~\cite{luo2018fast,liang2020pnpnet}. More recent methods, such as FutureDet~\cite{peri2022forecasting}, and DeTra~\cite{casas2024detra} address the joint detection and forecasting problem as a unified, single task. In camera-based systems, ViP3D~\cite{gu2023vip3d} leverages HD maps with tracking queries to capture object dynamics and semantics for forecasting, while follow-up works~\cite{hu2023planning, doll2024dualad, sun2024sparsedrive} extend this framework to planning with additional online map construction. More recently,  Foresight~\cite{papais2025foresight} builds on StreamPETR~\cite{wang2023exploring} and uses forward and backward query propagation to enhance both detection and forecasting performance.
In contrast, MASAR eliminates the need for tracking by constructing smooth past trajectories using appearance guidance and removes the reliance on map information by leveraging joint optimization of both detection and forecasting tasks.
We show though experiments substantial improvements over these SOTA models.

\section{METHOD} \label{method}
In the following sections, we describe our method for BEV-based models. Subsection~\ref{extension} then details how the same approach can be adapted for perspective-based models. 

\subsection{Overall Architecture} 

MASAR extends any transformer-based 3D object detector (DETR-like) into a joint object detector and forecaster. The APR module estimates smooth past trajectories for detected objects without explicit tracking, while the forecasting decoder leverages both object features and past trajectories to predict multi-modal future trajectories, capturing the long-term behavior of surrounding objects. An overview of the framework is shown in Figure~\ref{fig:model_architecture}.

\subsection{Scene Encoder} \label{sceneEncoder}

Given multi-frame, multi-view images, we use an image backbone~\cite{he2016deep} and a feature pyramid network (FPN)~\cite{lin2017feature} 
to extract multi-scale multi-view feature maps $\mathcal{F}_t$ for each time step $t$ in parallel, with $-T_{h} < t \le 0$,
where $T_h$ denotes the history length. For BEV-based detectors, the temporal multi-scale feature maps are fed to a BEV encoder to transform them into temporal BEV features~$\mathbf{B}_t$:
\begin{equation}
    \mathbf{B}_t = \text{BEVEncoder}(\mathcal{F}_t), \quad \mathbf{B}_t \in \mathbb{R}^{H_{\text{bev}} \times W_{\text{bev}} \times D}
\end{equation}

where $H_{\text{bev}}$ and $W_{\text{bev}}$ denote the spatial dimensions of the BEV grid, and $D$ is the feature dimension.

In our experiments, The BEVEncoder is a modification of the encoder proposed in~\cite{li2022bevformer}, composed of self-attention and spatial cross-attention layers. We remove the original auto-regressive temporal modeling and replace it with our object-centric spatio-temporal mechanism APR.

\subsection{Detector Decoder: Appearance-guided Past Motion Refinement (APR)} \label{detector}

The detector is a transformer-based decoder~\cite{li2022bevformer,liu2023sparsebev, wang2022detr3d} composed of $L_d$ refinement layers as shown in Figure~\ref{fig:model_architecture}.
It iteratively estimates how objects have moved over the past few seconds and leverages these trajectories to sample and aggregate corresponding features from previous frames \cite{lin2022sparse4d, liu2023sparsebev}. We formulate past trajectory generation as a candidate selection process: for each object, multiple trajectories are generated, and the most plausible one is selected based on a compatibility score computed from aggregated appearance features, which indicates how closely a trajectory matches the object’s true past motion, see Figure~\ref{fig:teaser}.

Algorithm~\ref{alg:apr} illustrates the iterative refinement process. 
With $N$ being the number of object queries, $M_h$ the number of past trajectory hypotheses, and $D$ the feature dimensionality, we denote object queries as $\mathbf{q}^{\text{obj}}_\ell \in \mathbb{R}^{N \times D}$, motion queries as $\mathbf{q}^{\text{mo}}_\ell \in \mathbb{R}^{N \times D}$, multi-hypothesis motion queries as $\mathbf{Q}^{\text{mo}}_\ell \in \mathbb{R}^{N \times M_h \times D}$, past trajectory proposals as $\mathbf{P}_\ell \in \mathbb{R}^{N \times T_h \times 2}$, and ego-motion–based transformation matrices $\{\mathbf{T}_{0 \to t}\}_{t=-1}^{1-T_h}$ for mapping positions from the current ego frame to past ego frames.

\begin{algorithm}[t]
\caption{Appearance-guided Past Motion Refinement}
\label{alg:apr}
\begin{algorithmic}[1]
\Require Temporal BEV $\mathbf{B}$, initial object queries $\mathbf{q}^{\text{obj}}_0$, motion emb. $\mathbf{E}_{\text{past}}$, trajectory proposals $\mathbf{P}_0$, and transforms $\{\mathbf{T}_{0 \to t}\}_{t=-1}^{1-T_h}$. 

\Ensure  Object queries $\mathbf{q}^{\text{obj}}$, motion queries $\mathbf{q}^{\text{mo}}$, and past trajectory proposals $\mathbf{P}$

\State $\mathbf{q}^{\text{mo}}_0 \gets \text{MLP}_{\text{init}}(\mathbf{q}^{\text{obj}}_0)$

\For{$\ell = 0$ to $L_d-1$}
    \State $\mathbf{Q}^{\text{mo}}_\ell \gets \mathbf{q}^{\text{mo}}_\ell + \mathbf{E}_{\text{past}}$ 
    \State $\mathbf{H}_\ell \gets \text{MotionDecoder}(\mathbf{Q}^{\text{mo}}_\ell, \mathbf{P}_\ell)$
    \State $\hat{\mathbf{H}}_\ell \gets \text{Concat}\{\mathbf{T}_{0 \to t} \cdot \mathbf{H}_\ell[:,:,t]\}_{t=-1}^{1-T_h}$
    \State $\mathbf{F}_\ell \gets \text{BEVSampler}(\mathbf{B}, \hat{\mathbf{H}}_\ell)$
    \State $\mathbf{F}_\ell^{\text{traj}} \gets \text{Aggregate}(\mathbf{F}_\ell,\mathbf{q}^{\text{obj}}_\ell)$
    \State $\mathbf{s}_\ell \gets \text{MLP}_{\text{score}}(\mathbf{F}_\ell^{\text{traj}})$
    \State $\mathbf{q}^{\text{obj}}_{\ell+1} \gets \mathbf{q}^{\text{obj}}_\ell + \sum_m \text{Softmax}(\mathbf{s}_\ell)_m \cdot \mathbf{F}_\ell^{\text{traj}}[:,m]$
    \State $m^\star \gets \arg\max_m \mathbf{s}_\ell$
    \State $\mathbf{q}^{\text{mo}}_{\ell+1} \gets \mathbf{Q}^{\text{mo}}_\ell[:, m^\star] + \text{MLP}_{\text{update}}(\mathbf{q}^{\text{obj}}_{\ell+1})$
    \State $\mathbf{P}_{\ell+1} \gets \mathbf{H}_\ell[:, m^\star, :, :2]$
\EndFor

\State \Return $\mathbf{q}^{\text{obj}}, \mathbf{q}^{\text{mo}}, \mathbf{P}$
\end{algorithmic}
\end{algorithm}

\paragraph*{\textbf{Initialization}} 
We initialize object queries $\mathbf{q}^{\text{obj}}_0$ following standard DETR-based 3D detection practices~\cite{wang2022detr3d, liu2023sparsebev}. Trajectory proposals $\mathbf{P}_0$ are initialized using a constant velocity model.
Motion queries $\mathbf{q}^{\text{mo}}_0$ are derived from object queries through an $\text{MLP}_{\text{init}}$ (line 1).

\paragraph*{\textbf{Past Motion Decoder} (lines 3-4)} 
We extend $\mathbf{q}^{\text{mo}}_\ell$ with a set of learnable embeddings $\mathbf{E}_{\text{past}} \in \mathbb{R}^{1 \times M_h \times D}$ to form trajectory hypothesis queries $\mathbf{Q}^{\text{mo}}_\ell$.
$\mathbf{E}_{\text{past}}$ acts as latent-space offsets, enabling the model to explore diverse variations of motion patterns efficiently.
We then predict multiple candidate past trajectories per object using $\mathbf{Q}^{\text{mo}}_\ell$ and trajectory proposals $\mathbf{P}_\ell$. The $\text{MotionDecoder}$ is similar to a PFD layer (Section~\ref{forecaster}) but employs only two factorized attention blocks: one over modes and one over time. Its output $\mathbf{H}_\ell \in \mathbb{R}^{N \times M_h \times T_h \times 4}$ encodes the means and scales of an isotropic Laplacian distributions for each time step and hypothesis, all expressed in the current $t=0$ ego-frame.

\paragraph*{\textbf{Temporal BEV Sampler} (lines 5-6)} 
We use multi-scale deformable attention~\cite{zhu2020deformable} to sample features from temporal BEVs along predicted past trajectory hypotheses. For each object and attention head $k$, a linear layer predicts $N_{\text{off}}$ sampling offsets $\Delta(x^k_i, y^k_i)$, scaled by the object bounding box dimensions $(w,l)$ similar to~\cite{lin2022sparse4d, liu2023sparsebev}. Another linear layer predicts attention weights $w^k_i$ from object queries $\mathbf{q}^{\text{obj}}_\ell$. Sampled features for head $k$ are computed as:
\begin{equation}
    \mathbf{f}^k_t = \sum_{i=1}^{N_{\text{off}}} w^k_i \, \text{Bilinear}\big(\mathbf{B}_t, \hat{\mathbf{H}}_\ell(t) + \Delta(x^k_i, y^k_i)\big),
\end{equation}

where $\hat{\mathbf{H}}_\ell$ denotes transformed trajectory hypotheses $\mathbf{H}_\ell$ to each previous frame ego coordinate system. Sampled features $\mathbf{F}_\ell \in \mathbb{R}^{N\times M_h \times T_h \times D}$ are obtained by concatenating $\mathbf{f}^k_t$ along the feature dimension and stacking across time.

\paragraph*{\textbf{Features Aggregator} (lines 7-9)}
To aggregate the sampled features~$\mathbf{F}_\ell$ of each object from temporal BEVs, we adopt adaptive mixing following~\cite{liu2023sparsebev, gao2022adamixer}, that we found provides both superior performance and greater flexibility compared to conventional temporal fusion strategies such as simple summation or recursive aggregation like in~\cite{lin2022sparse4d}. Subsequently, an $\text{MLP}_{\text{score}}$ predicts a compatibility score~$\mathbf{s}_\ell^{\text{traj}}$ based on the aggregated trajectory appearance features~$\mathbf{F}_\ell^{\text{traj}}$.
To update the object queries, we compute a weighted sum of the candidates' features~$\mathbf{F}_\ell^{\text{traj}}$, using the softmax-normalized scores as weights.  

\paragraph*{\textbf{Iterative Refinement} (lines 10-12)}
The trajectory proposals~$\mathbf{P}_{\ell+1}$ and motion queries~$\mathbf{q}^{\text{mo}}_{\ell+1}$ for the next decoder layer $\ell+1$ are obtained by selecting the highest-scoring hypothesis.

\subsection{Past-conditioned Forecasting Decoder (PFD)} \label{forecaster}

PFD is a transformer decoder where each block is composed of three factorized attention layers~\cite{zhou2023query, casas2024detra, ngiam2021scene} with a past-conditioning layer using cross-attention as illustrated in Figure~\ref{fig:model_architecture}. PFD refines a set of future queries $\mathbf{Q}^{\text{fut}}$, used to predict multi-modal future trajectories in a tracking-free and map-free setting. PFD is positioned hierarchically on top of a transformer-based object decoder forming a unified pipeline. The joint architecture is fully differentiable and trained end-to-end, enabling seamless optimization across both detection and forecasting tasks.

In contrast to prior works~\cite{papais2025foresight,casas2024detra,lang2024bevtp, hu2023planning},  PFD does not interact with any scene-level representation (e.g. BEV features or multi-view feature maps). We posit that object queries $\mathbf{q}^{\text{obj}}$ encode essential object semantics, while motion features $\mathbf{q}^{\text{mo}}$ and explicit past trajectories $\mathbf{P}$ provide sufficient information to capture object's dynamics.  

\paragraph*{\textbf{Query Initialization}}
We employ two sets of learnable parameters: $ \mathbf{e}^{\text{fut}}_t \in \mathbb{R}^d, \quad 1\le t <T_f$ for time steps embeddings and $\mathbf{e}_m^{\text{mode}} \in \mathbb{R}^d, \quad 1\le m <M_f$ for different future modes $M_f$~\cite{casas2024detra, zhou2023query}.  
These embeddings are expanded to construct a future query volume $\mathbf{Q}^{\text{fut}} \in \mathbb{R}^{N \times M_f \times T_f \times D}$.  

To make the future queries aware of the semantic features of each object, we incorporate information from the object queries $\mathbf{q}^{\text{obj}}$ using an MLP:  

\begin{equation}
    \mathbf{Q}^{\text{fut}} \leftarrow \text{LayerNorm}\big( \mathbf{Q}^{\text{fut}} + \text{MLP}(\mathbf{q}^{\text{obj}}) \big)
\end{equation}

\paragraph*{\textbf{Attention Layers}}
We employ four types of attention layers in our forecaster: past cross-attention, factorized temporal attention, factorized mode attention, and factorized social attention. 

For past-conditioning, the future queries $\mathbf{Q}^{\text{fut}}$ attend to past motion queries $\mathbf{q}^{\text{mo}}$ using cross-attention:
\begin{equation}
\begin{aligned}
    \tilde{\mathbf{Q}}^{\text{fut}} &\leftarrow 
    \text{CrossAttn}\big( \mathbf{Q}^{\text{fut}}, \mathbf{q}^{\text{mo}} + \text{PE}(\mathbf{P}) \big), \\
    \mathbf{Q}^{\text{fut}} &\leftarrow 
    \text{LayerNorm}\Big( \mathbf{Q}^{\text{fut}} + \tilde{\mathbf{Q}}^{\text{fut}} \Big),
\end{aligned}
\label{eq:fut_update}
\end{equation}
where $\text{PE}(.)$ denotes sinusoidal positional encoding.

 Factorized attention is widely adopted in motion prediction~\cite{sun2024sparsedrive,zhou2023query,ngiam2021scene}, as it is computationally efficient and captures rich dependencies across temporal, mode, and social dimensions in the query volume~\cite{casas2024detra}.  
For each factorized attention block, we first permute and reshape the query volume $\mathbf{Q}^{\textbf{fut}}$ so that the target factorized dimension (e.g., temporal $T_f$) is treated as the query dimension, while the remaining dimensions are flattened with the batch dimension~\cite{casas2024detra}. For instance, in the case of factorized temporal attention:
\begin{equation}
 \mathbf{Q}^{\text{fut}} \in \mathbb{R}^{N \times M_f \times T_f \times D}  \;\mapsto\; \tilde{\mathbf{Q}}^{\text{fut}} \in \mathbb{R}^{(N \cdot M_f) \times T_f \times D}   
\end{equation}
Self attention is then applied along $T_f$, and the result is reshaped back to recover the original query volume:
\begin{equation}
    \tilde{\mathbf{Q}}^{\text{fut}} \;\mapsto\; \mathbf{Q}^{\text{fut}} \in \mathbb{R}^{N \times M_f \times T_f \times D}
\end{equation}

Each attention layer is followed by a residual connection and layer normalization, and then passed by a position-wise feed-forward network~\cite{vaswani2017attention}. 

\begin{table*}[t]
\centering
\small
\setlength{\tabcolsep}{5pt}
\begin{tabular}{lc|cc|cc|ccccc}
\toprule
&  &\multicolumn{2}{c|}{Map} & 
\multicolumn{2}{c|}{3D Detection Metrics} & 
\multicolumn{4}{c}{Forecasting Metrics ($k=6$)} \\ 
\cmidrule(lr){3-4}\cmidrule(lr){5-6}\cmidrule(lr){7-10}
Method& Backbone & Offline & Online & mAP $\uparrow$ & NDS $\uparrow$ & EPA $\uparrow$ & minADE $\downarrow$ & minFDE $\downarrow$ & MR $\downarrow$ \\
\midrule
PnPNet~\cite{hu2023planning} & R101 & \cmark &  & -- &  --  & 0.222 & 1.15  & 1.95  & 0.226 \\   
PIP~\cite{jiang2022pip} & R50 &  & \cmark & 28.0 &  --  & 0.258 & 1.23  & 1.75  &  0.195 \\
UniAD~\cite{hu2023planning} & R101 &  & \cmark  & 38.0 & 49.8
& 0.456 &  0.71 & 1.02 & 0.151\\  
BEVFormer~\cite{li2022bevformer} + DeTra~\cite{casas2024detra}~\dag & R101 &  &  & 41.4 & --
& \underline{0.504} &  \underline{0.61} & \underline{1.00}& \underline{0.114}\\
\rowcolor{gray!20} BEVFormer~\cite{li2022bevformer} + Ours & R101 & & &
\textbf{43.0}& \textbf{52.9} & 
\textbf{0.519}&\textbf{0.55}&\textbf{0.82}&\textbf{0.101}\\
\midrule
Traditional~\cite{gu2023vip3d} & R50 & \cmark & &
--&--&
0.209&2.06&3.02&0.277\\
ViP3D~\cite{gu2023vip3d} & R50 & \cmark & &
--&--& 	
0.226&2.05&2.84&0.246\\
SparseDrive~\cite{sun2024sparsedrive} & R50 & & \cmark &
41.8 & 52.5& 	
0.482&\underline{0.62}&\underline{0.99}&\underline{0.136}\\
ForeSight~\cite{papais2025foresight} & R50 & \cmark & &
\textbf{46.6}& \textbf{56.0} & 	    
\textbf{0.499}&0.70&--&--\\
\rowcolor{gray!20} SparseBEV~\cite{liu2023sparsebev} + Ours & R50 & & &
\underline{43.5} & \underline{53.8} & 	
\underline{0.492}&\textbf{0.51}&\textbf{0.77}&\textbf{0.093}\\
\midrule
SparseDrive~\cite{sun2024sparsedrive} & R101 & & \cmark &
\underline{49.6}&\underline{58.8}&
\textbf{0.555}&\underline{0.60}&0.96&0.132\\
ForeSight~\cite{papais2025foresight} & R101 & \cmark & &
\textbf{50.2}&\textbf{58.9}&
0.549&0.68&\underline{0.93}&\underline{0.102}\\
\rowcolor{gray!20} SparseBEV~\cite{liu2023sparsebev} + Ours & R101 & & &
48.8&57.6&
0.544&\textbf{0.46}&\textbf{0.72}&\textbf{0.086}\\  
\bottomrule
\end{tabular}
\caption{Comparison of BEV-based and perspective-based methods on nuScenes joint detection and forecasting tasks. \dag~denotes DeTra’s refinement transformer integrated with BEVFormer (without map attention).}
\label{tab:sota_results}
\end{table*}

\paragraph*{\textbf{Multi-Modal Futures Regression and Scoring}}

At the end of each forecaster layer, we employ a regression head that for each object $i$ and each future mode $m$, predicts the future location means $\mu^{i,m}_t \in \mathbb{R}^2$ and scales $\sigma^{i,m}_t \in \mathbb{R}^2$ of an isotropic Laplacian distribution~\cite{zhou2022hivt}. We use another head to predict the scores associated with each future mode. 

\subsection{Adaptation to Perspective-based Models} \label{extension}
For perspective-based models such as SparseBEV~\cite{liu2023sparsebev}, we remove the BEV encoder and allow the detector decoder to directly interact with temporal multi-view features~$\mathcal{F}_t$. We assume that object motion occurs only in the BEV plane~\cite{liu2023sparsebev}, keeping the same $z$ coordinate along the past trajectory. For sampling, we use the same mechanism introduced in~\cite{liu2023sparsebev}. PFD remains unchanged.

\section{EXPERIMENTS} \label{experiments}
In this section, we detail the dataset, evaluation metrics, and implementation specifics used to assess MASAR's performance on joint detection and forecasting.
\subsection{Datasets}
To compare against most prior works on vision-based joint detection and forecasting, we train and evaluate our framework on nuScenes, a large-scale autonomous driving dataset~\cite{caesar2020nuscenes}.
It contains 1000 driving scenes, each lasting approximately 20 seconds. The dataset includes annotations for 3D detection and tracking at 2\,Hz, which we use to construct ground-truth trajectories for end-to-end forecasting~\cite{hu2023planning}. All results are reported on the validation split.

\subsection{Evaluation Metrics}  
For 3D detection, we follow the official nuScenes 3D detection benchmark~\cite{caesar2020nuscenes} and report main metrics, including mean Average Precision (mAP) and NuScenes detection score (NDS).  
For trajectory forecasting, we follow the evaluation protocol from~\cite{hu2023planning}, reporting metrics for $T_f = 12$ time steps, and over $M_f = 6$ predicted future modes, including minADE, minFDE, and Miss Rate. These metrics are computed on matched objects at 2.0 meters threshold and averaged similarly to TP metrics in detection.  
Additionally, we report the End-to-End Prediction Accuracy (EPA)~\cite{gu2023vip3d}, which accounts for both false positive detections and successful forecasts (i.e., true positive detections with an FDE less than 2 meters).

\subsection{Implementation Details}  
As highlighted in Section~\ref{method}, our framework is flexible and can be integrated with both BEV-based and perspective-based object detectors. Accordingly, our experiments build upon two models: \textbf{BEVFormer}~\cite{li2022bevformer} a BEV-based 3D detector and \textbf{SparseBEV}~\cite{liu2023sparsebev} a perspective-based 3D detector. Unless otherwise specified, we follow the official training settings for each model, including image backbone, input resolution, number of object queries, optimizer, and detection loss $\mathcal{L}_{\text{det}}$. 

The overall training objective for the joint detection, past refinement and forecasting is:

\begin{equation}
\mathcal{L} = \mathcal{L}_{\text{det}} + \lambda_p \cdot \mathcal{L}_{\text{past}} + \lambda_f \cdot \mathcal{L}_{\text{future}} \,
\end{equation}

Our newly introduced trajectory-related losses $\mathcal{L}_{\text{past}}$ and $\mathcal{L}_{\text{future}}$ consist each of two components:
(i) a regression loss, computed as the negative log-likelihood of an isotropic Laplacian distribution~\cite{zhou2022hivt}, and  
(ii) a scoring loss. For past refinement, the scoring loss is formulated as a binary cross-entropy loss, where targets are derived by scaling the negative average displacement error into $[0,1]$ via a sigmoid function. For future prediction, the scoring loss is supervised using cross-entropy with a soft-target strategy~\cite{zhou2022hivt}. We compute $\mathcal{L}_\text{past}$ and $\mathcal{L}_\text{future}$ only for assigned objects with~$\text{center distance} \le 1.0\text{ m}$ to the ground truth~\cite{casas2024detra}. We adopt a teacher forcing with curriculum schedule training strategy: ground-truth past trajectories are used in place of the refined trajectories, and their ratio is gradually decreased over time.

The full model is trained end-to-end in a fully differentiable manner. We use $\lambda_p = 0.2$ and $\lambda_f = 0.1$ for weighting the past and future trajectory losses. 
\subsection{Joint Detection and Forecasting Performance} 

We evaluate our joint detection and forecasting framework against state-of-the-art models on the nuScenes validation split~\cite{caesar2020nuscenes}. Since BEV-based methods typically fall behind perspective-based methods in 3D detection~\cite{li2024bevnext}, we group comparisons into two categories: BEV-based and perspective-based approaches. Within perspective-based models, we further group methods according to the image backbone used.

\subsubsection{BEV-based Models} As shown in Table~\ref{tab:sota_results}, our BEVFormer~\cite{li2022bevformer} variant equipped with both APR and PFD achieves the best performance on end-to-end forecasting metrics, reducing minFDE by $17.6\%$ and improving EPA by $12.5\%$ compared to prior work~\cite{hu2023planning}. Furthermore, our model significantly enhances BEVFormer~\cite{li2022bevformer} in 3D detection (3.3\% improvement in mAP). Table~\ref{tab:results_per_cls} reports detailed forecasting metrics for cars and pedestrians. Our method improves forecasting for dynamic classes, achieving nearly $13\%$ improvement in minADE for cars and $6\%$ for pedestrians compared to the baseline (BEVFormer + DeTra~\cite{casas2024detra}).

\subsubsection{Perspective-based Models} Our integration with SparseBEV~\cite{liu2023sparsebev} surpasses all prior methods in motion prediction, achieving $20\%$ lower minADE and minFDE and a $13\%$ lower miss rate compared to~\cite{sun2024sparsedrive, papais2025foresight}, all without using map information during training or inference. Although our model is not the top ranked model in 3D detection, it attains highly competitive EPA (within $1\%$ of the best), demonstrating the effectiveness of our past-conditioned forecaster in modeling diverse future agent behaviors, which in turn leads to higher true positive joint detections and forecasts~\cite{gu2023vip3d}. As shown in Table~\ref{tab:results_per_cls}, MASAR achieves the best performance for pedestrians, surpassing DualAD~\cite{doll2024dualad} by 8\% in EPA and 4\% in minADE, highlighting its ability to model complex agent behaviors. 

\begin{table}[thb]
\centering
\renewcommand{\arraystretch}{1.2} 
\setlength{\tabcolsep}{3pt} 
\begin{tabular}{l|cc >{\columncolor{gray!20}}c |cc|cc}
\toprule 
\textbf{Method} & \multicolumn{3}{c|}{\textbf{EPA}$\uparrow$} & \multicolumn{2}{c|}{\textbf{minADE}$\downarrow$} & \multicolumn{2}{c}{\textbf{minFDE}$\downarrow$} \\
 &Car&Ped&Avg&Car&Ped&Car&Ped\\ 
\midrule 
UniAD~\cite{hu2023planning}& 0.46&0.35&0.405 &\textbf{0.71}&0.78 & \textbf{1.02} & 1.05 \\ 
BEVFormer~\cite{li2022bevformer} + DeTra&0.49&0.43&0.46&0.84&0.74&1.61 & 1.12\\ 
BEVFormer~\cite{li2022bevformer} + Ours&\textbf{0.50}&\textbf{0.45}&\textbf{0.475}&0.73&\textbf{0.69} & 1.19 & \textbf{1.00} \\ 
\midrule 
SparseDrive-S~\cite{sun2024sparsedrive}&0.48 & 0.41 & 0.445& 0.62 & 0.72& \textbf{0.99}& 1.07 \\
DualAD~\cite{doll2024dualad}&\textbf{0.524}&0.452&0.488&0.68&0.63&1.08&0.89\\ 
SparseBEV~\cite{liu2023sparsebev} + Ours&0.522&\textbf{0.491}&\textbf{0.506}&\textbf{0.60}&\textbf{0.60}&1.08& \textbf{0.88}\\
\bottomrule
\end{tabular}
\caption{Comparison of motion prediction results on dynamic classes (cars and pedestrians). Results for ForeSight~\cite{papais2025foresight} are not reported, as per-class metrics are not publicly available.}
\label{tab:results_per_cls}
\end{table}

\subsection{Ablation Studies} 
We conduct several ablation experiments to assess the impact of each component on MASAR's overall performance. 
\subsubsection{Impact of History Length} 
To examine the impact of longer historical context on joint detection and forecasting, we vary the number of past frames provided to BEVFormer-small~\cite{li2022bevformer}. As shown in Table~\ref{tab:bevformer_ablations},  increasing past frames to 8 (4 seconds) steadily improves detection and forecasting metrics ($8.5\%$ improvement on NDS). While it is shown that BEVFormer's recurrent temporal modeling saturates around 4 frames~\cite{li2022bevformer,han2024exploring}, APR effectively leverages longer histories, yielding consistent gains.

\begin{table}[thb]
\centering
\renewcommand{\arraystretch}{1.2}
\setlength{\tabcolsep}{4pt}
\begin{tabular}{cccccccc}
\toprule
\textbf{Frames} & \textbf{NDS} $\uparrow$ & \textbf{mAP} $\uparrow$ & \textbf{EPA} $\uparrow$ & \textbf{minADE} $\downarrow$ & \textbf{minFDE} $\downarrow$ & \textbf{MR} $\downarrow$ \\
\midrule
\rowcolor{gray!20} 3 & 47.8 & 37.0 & -- & -- & -- & -- \\  
4 & 50.7 & 40.0 & 0.491 & 0.592 & 0.868 & 0.108 \\
6 & 51.4 & 40.8 & 0.498 & 0.587 & 0.862 & 0.105 \\
8 & \textbf{51.9} & \textbf{41.5} & \textbf{0.500} & \textbf{0.583} & \textbf{0.856} & \textbf{0.104} \\
\bottomrule
\end{tabular}
\caption{Detection and forecasting performance with varying number of past frames on BEVFormer-small~\cite{li2022bevformer}. The gray row indicates the original BEVFormer-small performance taken from official BEVFormer Github repository.}
\label{tab:bevformer_ablations}
\end{table}

\subsubsection{Past-conditioned Forecasting Decoder}
Table~\ref{tab:ablation_components_pfd} evaluates the contribution of each component in the forecasting decoder. We conduct this experiment using SparseBEV~\cite{liu2023sparsebev} with a R50 image backbone. Removing object queries from the future query initialization results in the loss of semantic and class information about the objects, while omitting the learnable time and mode weights limits the model’s ability to generate distinct queries for different future possibilities. Additionally, past conditioning significantly improves forecasting performance, confirming our first motivation of ``looking backward to look forward''. 

\begin{table}[thb]
\centering
\renewcommand{\arraystretch}{1.2} 
\setlength{\tabcolsep}{5pt}       
\begin{tabular}{ccc cccc}
\toprule
\textbf{W} & \textbf{O} & \textbf{P} & \textbf{EPA}$\uparrow$ & \textbf{minADE}$\downarrow$ & \textbf{minFDE}$\downarrow$ & \textbf{MR}$\downarrow$ \\
\midrule
\xmark & \cmark & \cmark & 0.432 & 1.064 & 1.786 & 0.194 \\
\cmark & \xmark & \cmark & 0.480 & 0.594 & 0.913 & 0.106 \\
\cmark & \cmark & \xmark & 0.489 & 0.539 & 0.820 & 0.100 \\
\cmark & \cmark & \cmark & \textbf{0.492} & \textbf{0.511} & \textbf{0.770} & \textbf{0.093} \\
\bottomrule
\end{tabular}
\caption{Ablation of forecasting decoder components.\\ 
W: learnable weights for query initialization, O: object queries, P: past cross-attention.}
\label{tab:ablation_components_pfd}
\end{table}

\subsubsection{Past Motion Modeling} 
We evaluate three past motion modeling variants using SparseBEV~\cite{liu2023sparsebev} with an R50 backbone: constant velocity, past motion prediction without appearance guidance (w/o AG), and with appearance guidance (w/ AG). As shown in Table~\ref{tab:ablation_components_apr}, past refinement with AG slightly reduces detection but improves forecasting, achieving a 6\% gain in minFDE. This highlights the challenge of jointly performing detection and past predictions. While many objects in nuScenes~\cite{caesar2020nuscenes} are static—making constant velocity sufficient for detection—appearance guidance better captures past object dynamics, enhancing forecasting performance while maintaining robust detection compared to not using guidance.
\begin{table}[H]
\centering
\renewcommand{\arraystretch}{1.2} 
\setlength{\tabcolsep}{3pt} 
\begin{tabular}{lcccccc}
\toprule 
\textbf{Past Motion} &\textbf{mAP}$\uparrow$ &\textbf{EPA}$\uparrow$ & \textbf{FDE}$_{\text{past}}$ $\downarrow$ &\textbf{minADE}$\downarrow$ & \textbf{minFDE}$\downarrow$ & \textbf{MR}$\downarrow$ \\
\midrule 
Const. Vel. & \textbf{44.2} & \textbf{0.49} & 0.97 & 0.53 & 0.82 & 0.104 \\
w/o AG & 42.8 & 0.48 & \underline{0.84} &\underline{0.52} & \underline{0.78} & \underline{0.095} \\
w/ AG & \underline{43.5} & \textbf{0.49} & \textbf{0.83}  & \textbf{0.51} & \textbf{0.77} & \textbf{0.093} \\
\bottomrule
\end{tabular}
\caption{Ablation of past motion modeling strategies. $\text{FDE}_{\text{past}}$ denotes single hypothesis FDE computed for past motion on dynamic classes (cars and pedestrians).} 
\label{tab:ablation_components_apr}
\end{table}

\begin{figure*}[t]
    \centering
    \includegraphics[width=0.98\textwidth]{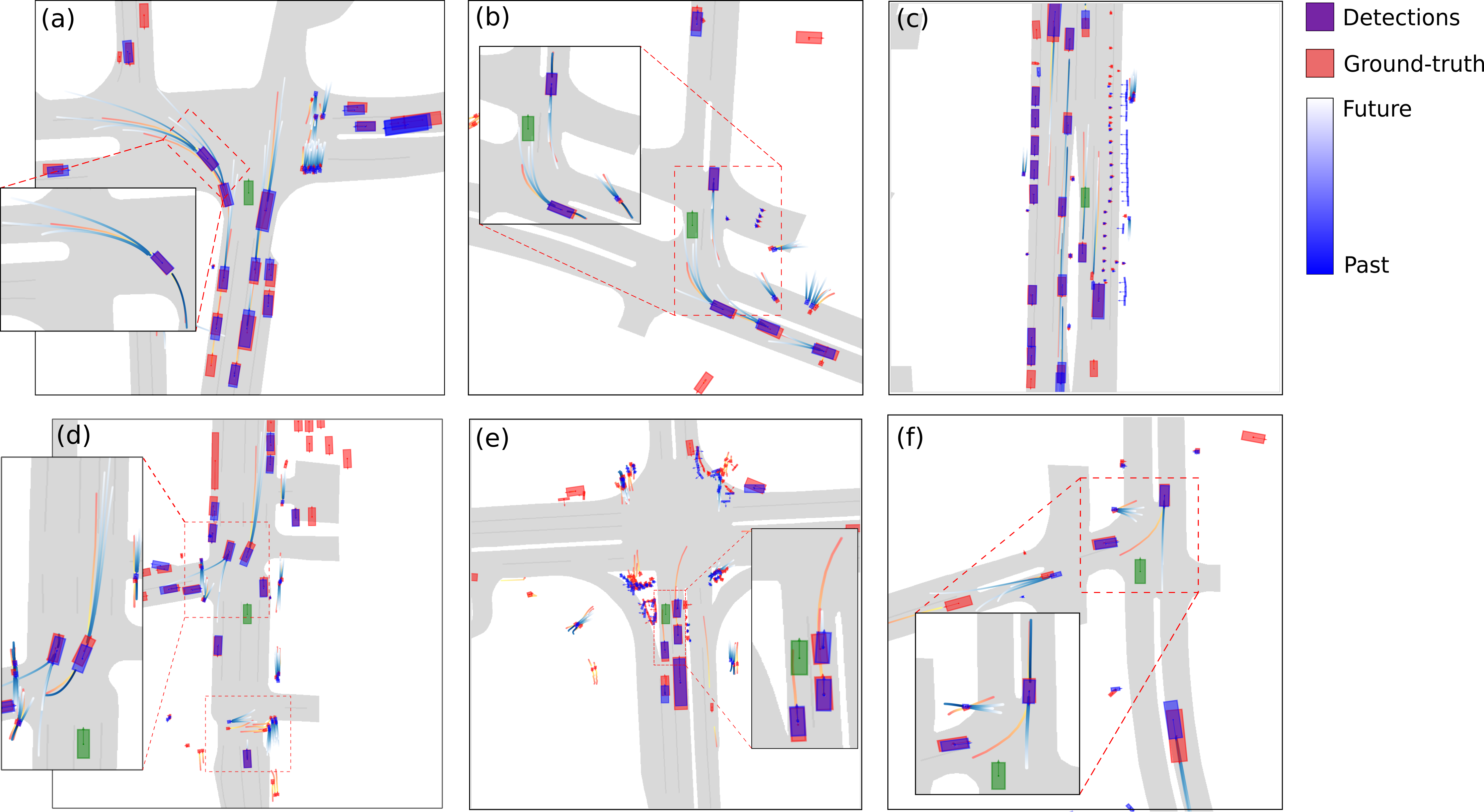} 
    \caption{Visualization of some nuScenes validation samples. Green: ego, purple: detections, red: ground-truth. Only future trajectories are plotted; past trajectories are shown in magnified views. The rendered maps are just for visualization and not input to the model. (a) and (c) show challenging crowded scenes, (b) and (d) diverse multi-modal futures, and (e) and (f) typical failure cases from missing context.}
    \label{fig:vis_samples}
\end{figure*}

\subsection{Qualitative results} 
Figure~\ref{fig:vis_samples} shows selected qualitative examples from nuScenes. Samples (a) and (c) highlight MASAR’s ability to generate diverse and consistent future trajectories for different agent types, including cars, pedestrians, motorcycles, and large vehicles. Samples (b) and (d) demonstrate MASAR’s capacity to capture plausible multi-modal futures; in (d), for instance, the model correctly predicts pedestrians’ intention to cross the street, although it fails to capture a change of direction for another group of pedestrians at the bottom of the figure. In contrast, (e) shows difficulty in handling behavior changes of stopped vehicles, while (f) illustrates a failure case where the model correctly estimates an oncoming car’s past trajectory but misses its future intention, suggesting the potential benefit of incorporating HD map context.

\section{CONCLUSIONS}
In this work, we introduced MASAR, a unified framework for joint 3D detection and trajectory forecasting in autonomous driving. We designed a new object-centric spatio-temporal mechanism leveraging motion and appearance cues by refining past trajectories with appearance guidance. MASAR captures long-term temporal dependencies without relying on tracking or any map information. We integrated it with transformer-based detectors such as BEVFormer and SparseBEV, showing consistent improvements over prior end-to-end methods in trajectory forecasting. Extensive ablation studies demonstrate the effectiveness of each component in our framework, highlighting how past-conditioning substantially improves future forecasting.





\balance
\bibliographystyle{IEEEtran} 
\bibliography{refs}

\end{document}